\documentclass[10pt,twocolumn,letterpaper]{article}

\usepackage{cvpr}
\usepackage{graphicx}
\usepackage{amsmath,amssymb} 
\usepackage{color}

\DeclareMathOperator*{\argmin}{argmin} 
\DeclareMathOperator*{\argmax}{argmax}
\usepackage[font=scriptsize]{caption}
\usepackage{booktabs}
\usepackage{array}
\usepackage[breaklinks=true,bookmarks=false]{hyperref}
\newcolumntype{C}[1]{>{\centering\let\newline\\\arraybackslash\hspace{0pt}}m{#1}}
\usepackage{times,txfonts}
\usepackage[caption = false]{subfig}
\usepackage{caption}
\usepackage{url}

\cvprfinalcopy
\begin{document}

\title{A Recurrent Encoder-Decoder Network for Sequential Face Alignment} 





\author{Xi Peng\\
Rutgers University\\
{\tt\small xipeng.cs@rutgers.edu}
\and
Rogerio Feris\\
IBM T. J. Watson\\
{\tt\small rsferis@us.ibm.com}
\and
Xiaoyu Wang\\
Snapchat Research\\
{\tt\small fanghuaxue@gmail.com}
\and
Dimitris Metaxas\\
Rutgers University\\
{\tt\small dnm@cs.rutgers.edu}
}

\maketitle

\begin{abstract}
We propose a novel recurrent encoder-decoder network model for real-time video-based face alignment. Our proposed model predicts 2D facial point maps regularized by a regression loss, while uniquely exploiting recurrent learning at both spatial and temporal dimensions. At the spatial level, we add a feedback loop connection between the combined output response map and the input, in order to enable iterative coarse-to-fine face alignment using a {\em single network model}. At the temporal level, we first decouple the features in the bottleneck of the network into {\em temporal-variant factors}, such as pose and expression, and {\em temporal-invariant factors}, such as identity information. Temporal recurrent learning is then applied to the decoupled temporal-variant features, yielding better generalization and significantly more accurate results at test time. We perform a comprehensive experimental analysis, showing the importance of each component of our proposed model, as well as superior results over the state-of-the-art in standard datasets.
\end{abstract}

\section{Introduction}
Face landmark detection plays a fundamental role in many computer vision tasks, such as face recognition, expression analysis, and 3D face modeling. In the past few years, many methods have been proposed to address this problem, with significant progress being made towards systems that work in real-world conditions (``in the wild'').

Regression-based approaches \cite{XiongCVPR13,CaoIJCV14} have achieved impressive results by cascading discriminative regression functions that directly map facial appearance to landmark coordinates. In this framework, deep convolutional neural networks have proven effective as a choice for feature extraction and non-linear regression modeling \cite{ZhangECCV14,Lai2015,ZhangTangECCV14}. Although these methods can achieve very reliable results in standard benchmark datasets, they still suffer from limited performance in challenging scenarios, e.g., involving large face pose variations and heavy occlusions.

A promising direction to address these challenges is to consider video-based face alignment (i.e., sequential face landmark detection) \cite{ShenICCVW15}, leveraging temporal information as an additional constraint~\cite{WangCVPR16}. Despite the long history of research in rigid and non-rigid face tracking \cite{BlackCVPR95,OliverCVPR97,DecarloIJCV00,PatrasFG04}, current efforts have mostly focused on face alignment in still images \cite{SagonasICCVW13,ZhangECCV14,TzimiropoulosCVPR15,ZhuCVPR15}. In fact, most methods often perform video-based landmark detection by independently applying models trained on still images in each frame in a tracking-by-detection manner \cite{WangTPAMI15}, with notable exceptions such as \cite{AsthanaCVPR14,PengICCV15}, which explore incremental learning based on previous frames. How to effectively model long-term temporal constraints while handling large face pose variations and occlusions is an open research problem for video-based face alignment.

In this work, we address this problem by proposing a novel recurrent encoder-decoder deep neural network model (see Figure \ref{fig:fig_overview}). The encoding module projects image pixels into a low-dimensional feature space, whereas the decoding module maps features in this space to 2D facial point maps, which are further regularized by a regression loss. In order to handle large face pose variations, we introduce a feedback loop connection between the aggregated 2D facial point maps and the input. The intuition is similar to cascading multiple regression functions \cite{XiongCVPR13,ZhangECCV14} for iterative course-to-fine face alignment, but in our approach the iterations are modeled jointly with shared parameters, using a single network model. 

For more effective temporal modeling, we first decouple the features in the bottleneck of the network into temporal-variant factors, such as pose and expression, and temporal-invariant factors, such as identity. More specifically, we split the features into two components, where one component is used to learn face recognition using identity labels, and recurrent temporal learning is applied to the other component, which encodes temporal-variant factors only. We show in our experiments that recurrent learning in both spatial and temporal dimensions is crucial to improve performance of sequential face landmark detection. In summary, our work makes the following {\bf contributions}:

\begin{itemize}
\item We propose a novel recurrent encoder-decoder network model for real-time sequential face landmark detection. To the best of our knowledge, this is the first time a recurrent model is investigated to perform video-based facial landmark detection.
\item Our proposed {\em spatial recurrent learning} enables a novel iterative coarse-to-fine face alignment using a single network model. This is critical to handle large face pose changes and a more effective alternative than cascading multiple network models in terms of accuracy and memory footprint.
\item Different from traditional methods, we apply {\em temporal recurrent learning} to temporal-variant features which are decoupled from temporal-invariant features in the bottleneck of the network, achieving better generalization and more accurate results. 
\item We provide a detailed experimental analysis of each component of our model, as well as insights about key contributing factors to achieve superior performance over the state-of-the-art.  The project page is public available. \footnote{\url{https://sites.google.com/site/xipengcshomepage/project/face-alignment}}
\end{itemize}

\section{Related Work}
Face alignment has been advanced in last decades. Remarkably, regression based methods~\cite{AsthanaCVPR13,SunCVPR13,XiongCVPR13,CaoIJCV14,ZhangECCV14,AsthanaCVPR14,ZhuCVPR15,TzimiropoulosCVPR15,JourablooCVPR16,WuCVPR16,ZhuCVPR16} significantly boost the generalization performance of face landmark detection, compared to algorithms based on statistical models such as Active shape models \cite{Cootes92BMVC,StephenECCV08} and Active appearance models~\cite{Gao2010}. 
A regression-based approach directly regresses landmark locations where features extracted from face images serve as regressors. Landmark models are learned either in an independent manner, or in a joint fashion \cite{CaoIJCV14}.
This paper performs landmark detection via both a classification model and a regression model. Different from most of the previous methods, this work deals with face alignment in a video. It jointly optimizes detection output by utilizing multiple observations from the same person.

Learning cascade-like regression models show superior performance on the face alignment task~\cite{XiongCVPR13,SunCVPR13,ZhangECCV14}. Supervised descent method~\cite{XiongCVPR13} learns cascades of regression models based on SIFT feature.
Sun \emph{et. al.}~\cite{SunCVPR13} proposed to use three levels of neural networks to predict landmark locations. 
Zhang \emph{et. al.}~\cite{ZhangECCV14} studied the problem via cascades of stacked auto-encoders which gradually refine the landmark position with higher resolution inputs. 
Compared to these efforts which explicitly define cascade structures, our method learns a spatial recurrent model which implicitly incorporates the cascade structure with shared parameters. It is also more "end-to-end" compared to previous works that handcraftly divide the learning process into multiple stages. 

Recurrent neural networks (RNNs) are widely employed in the literature of speech recognition~\cite{MikolovInterspeech10} and natural language processing~\cite{MikolovArxiv14}. 
They are also recently used in computer vision. For example, in the task of image captioning~\cite{Karpathy_2015_CVPR} and video captioning~\cite{Yao_2015_ICCV}, RNNs are employed for text generation. Veeriah \emph{et. al.}~\cite{VeeriahICCV15} use RNNs to learn complex time-series representations via high-order derivatives of states for action recognition. Benefiting from the deep architecture, RNNs are naturally good alternatives to Conditional Random Fields (CRFs) ~\cite{ZhengICCV15} which are popular in image segmentation. 

Encoder and decoder networks are well studied in machine translation~\cite{ChoArxiv14} where the encoder learns the intermediate representation and the decoder generates the translation from the representation. It is also investigated in speech recognition~\cite{llu_is2015b} and computer vision~\cite{BadriCoRR15,HongCoRR15}. Yang \emph{et. al.}~\cite{YangNIPS15} proposed to decouple identity units and pose units in the bottleneck of the network for 3D view synthesis. However, how to fully utilize the decoupled units for correspondence regularization \cite{LongNIPS14} is still unexplored. In this work, we employ the encoder to learn a joint representation for identity, pose, expression as well as landmarks. The decoder translates the representation to landmark heatmaps. Our spatial recurrent model loops the whole encoder-decoder framework.

\section{Recurrent Encoder-Decoder Network}
In this section, we first give an overview of our approach. Then we describe the novelty of our work in detail: spatial and temporal recurrent learning, supervised identity disentangling, and constrained shape prediction. 

\subsection{Method Overview}
\begin{figure*}
\centering
\includegraphics[width=1\textwidth]{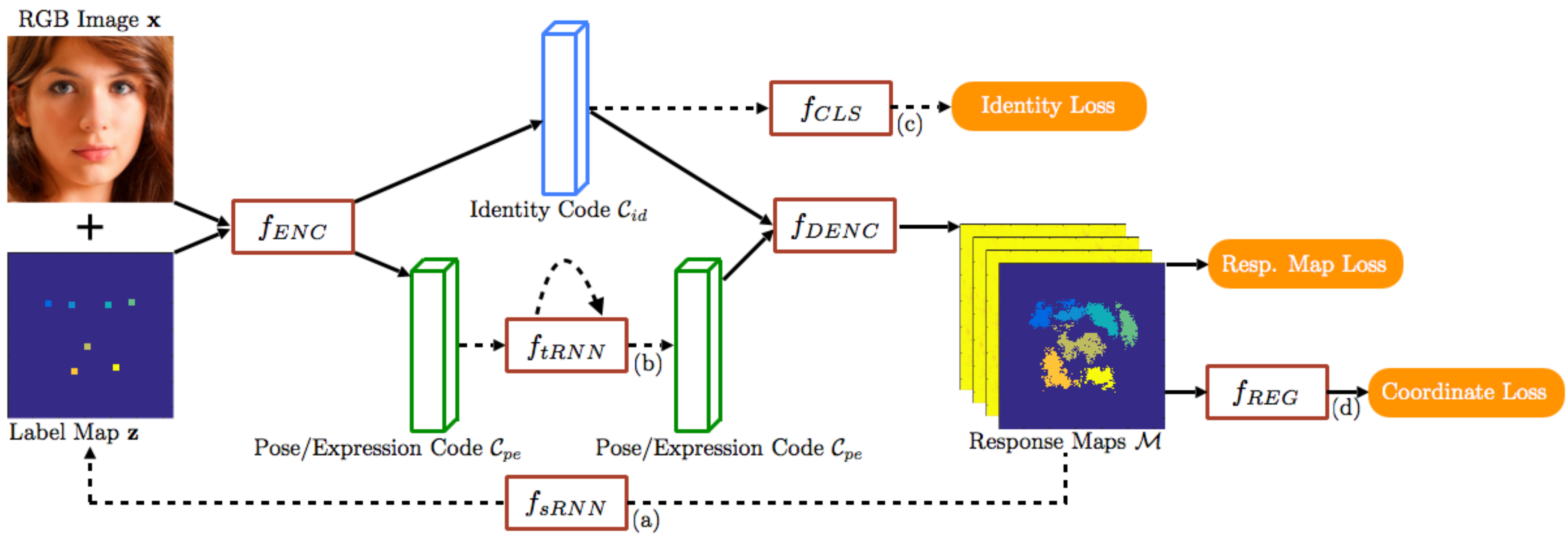}
\caption{Overview of the recurrent encoder-decoder network: \textbf{(a)} spatial recurrent learning (Section \ref{sec:spatial}); \textbf{(b)} temporal recurrent learning (Section \ref{sec:temporal}); \textbf{(c)} supervised identity disentangling (Section \ref{sec:identity}); and \textbf{(d)} constrained shape prediction (Section \ref{sec:shape}). $f_{ENC},f_{DENC},f_{sRNN},f_{tRNN},f_{CLS},f_{REG}$ are potentially nonlinear and multi-layered mappings.}
\label{fig:fig_overview}
\end{figure*}

Our task is to locate $L$ landmarks in sequential images using an end-to-end deep neural network. Figure \ref{fig:fig_overview} shows the overview of our approach. We consider $f_{\star}$ as potential nonlinear and multi-layered functions. The input of the network are the image $\mathbf{x} \in \mathbb{R}^{w \times h \times 3}$ and the landmark label map $\mathbf{z} \in \mathbb{R}^{w \times h \times 1}$. Each pixel in $\mathbf{z}$ is a discrete label $\{0,\cdots,L\}$ that marks the presence of the corresponding landmark, where $0$ denotes a non-landmark area.

The \textit{encoder} ($f_{ENC}$) performs a sequence of convolution, pooling and batch normalization \cite{IoffeCoRR15} to extract a representation code from inputs:
\begin{align}
 	\mathcal{C} = f_{ENC}(\mathbf{x},\mathbf{z}; \theta_{ENC}), \; \mathcal{C} \in \mathbb{R}^{w_c \times h_c \times d_c},
\end{align}
where $\mathcal{C}$ represents the encoded features. $\theta_{ENC}$ denotes encoder parameters. Symmetrically, the \textit{decoder} ($f_{DENC}$) performs a sequence of unpooling, convolution and batch normalization to upsample the representation codes to a multi-channel response map:
\begin{align}
	\mathcal{M} = f_{DENC}(\mathcal{C}; \theta_{DENC}), \; \mathcal{M} \in \mathbb{R}^{w \times h \times (L+1)},
\end{align}
where $\theta_{DENC}$ denotes the decoder parameters. The first channel of $\mathcal{M}$ represents the background, while the rest $L$ channels of $\mathcal{M}$ present pixel-wise confidence of the corresponding landmarks. The $(L+1)$-channel response map is crucial to preserve the landmark unity, compared with a 2-channel setup (landmark {\em v.s.} non-landmark).

The encoder-decoder framework plays an important role in our task. {\bf First}, it is convenient to perform {\em spatial recurrent learning} ($f_{sRNN}$) since $\mathcal{M}$ has the same dimension (but different number of channels) as $\mathbf{x}$. The output of the decoder can be directly fed back into the encoder to provide pixel-wise spatial cues for the next recurrent step. {\bf Second}, we can decouple $\mathcal{C}$ in the bottleneck of the network into temporal-variant and -invariant factors. The former is further exploited in {\em temporal recurrent learning} ($f_{tRNN}$) for robust alignment, while the latter  is used in {\em supervised identity disentangling} ($f_{CLS}$) to facilitate the network training. {\bf Third}, $\mathcal{M}$ can be further regularized in {\em constrained shape prediction} ($f_{REG}$) to directly output landmark coordinates. The details of each module are explained in following subsections.

\begin{figure}[t]
\minipage{0.49\textwidth}
  \includegraphics[width=0.89\linewidth]{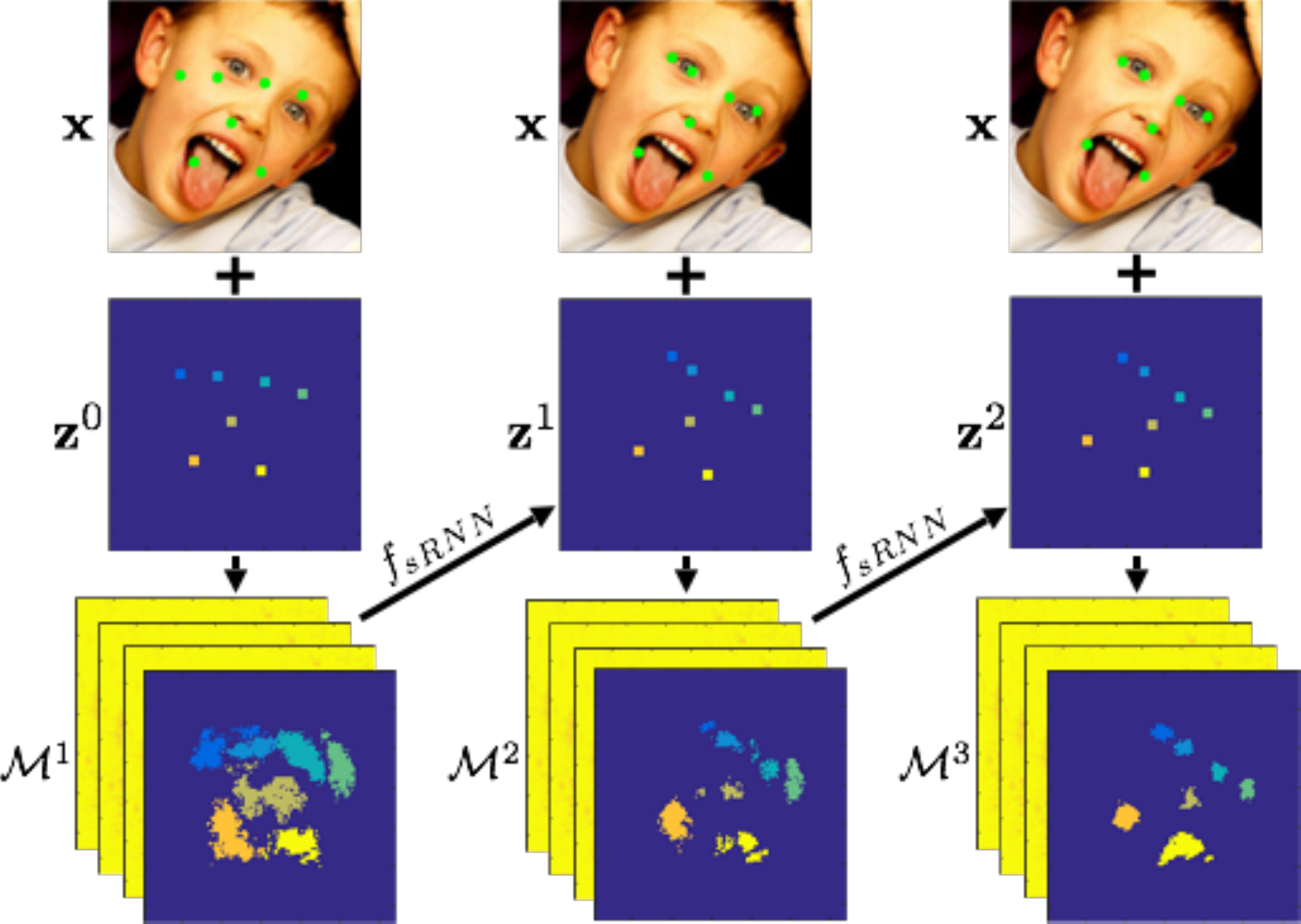}
  \caption{An unrolled illustration of {\em spatial recurrent learning}. The response map is pretty coarse when the initial guess is far away from the ground truth if large pose and expression exist. It eventually gets refined in the successive recurrent steps.}\label{fig:fig_fsrnn}
\endminipage\hfill
\minipage{0.49\textwidth}
  \includegraphics[width=\linewidth]{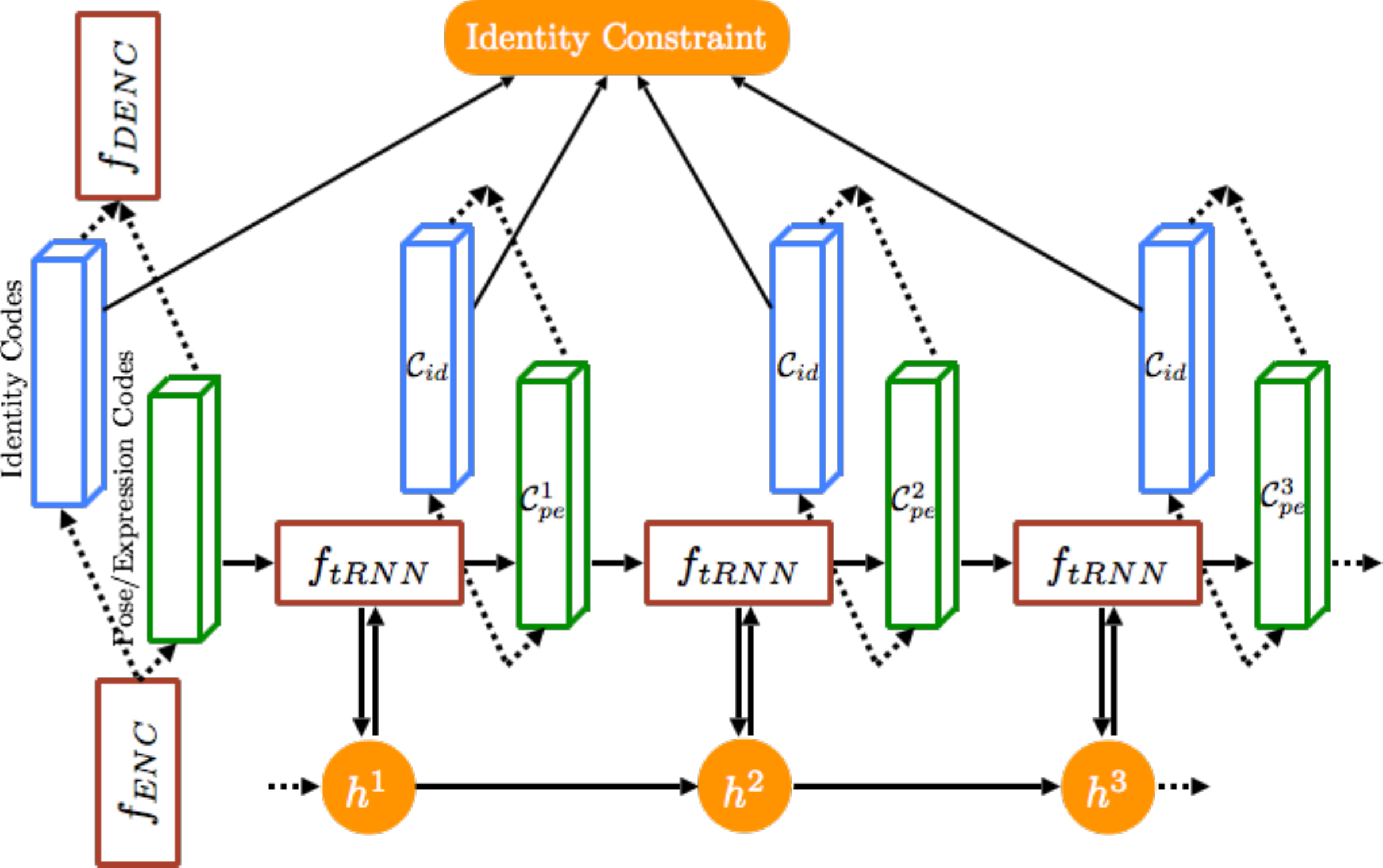}
  \caption{An unrolled illustration of {\em temporal recurrent learning}. $\mathcal{C}_{id}$ encodes temporal-invariant factor which subjects to the same identity constraint. $\mathcal{C}_{pe}$ encodes temporal-variant factors which is further modeled in $f_{tRNN}$.}\label{fig:fig_ftrnn}
\endminipage\hfill
\end{figure}

\subsection{Spatial Recurrent Learning} \label{sec:spatial}
The purpose of spatial recurrent learning is to pinpoint landmark locations in a coarse-to-fine manner. Unlike existing approaches \cite{SunCVPR13,ZhangECCV14} that employ multiple networks in cascade, we accomplish the coarse-to-fine search in a single network in which the parameters are jointly learned in successive recurrent steps.

Given an image $\mathbf{x}$ and initial guess of the shape $\mathbf{z}_0$, we refine the shape prediction iteratively $\{\mathbf{z}^1,\cdots,\mathbf{z}^k\}$ by feeding back the previous prediction:
\begin{align}
	\mathbf{z}^k = f_{sRNN}(\mathcal{M}^{k-1}) = f_{sRNN}( f_{DENC}( f_{ENC}(\mathbf{x},\mathbf{z}^{k-1}) ) ), \; k=1,\cdots,K,
\end{align}
where we omit network parameters $\theta_{ENC}$ and $\theta_{DENC}$ for concise expression. The network parameters are learned by recurrently minimizing the classification loss between the annotation and the response map output by the encoder-decoder:
\begin{align} \label{eq:srnnloss}
	\argmin_{\theta_{ENC},\theta_{DENC}} \sum_{k=1}^{K} \sum_{l=0}^{L} \ell ( \mathcal{M}^*_l, f_{DENC}( f_{ENC}(\mathbf{x}, \mathbf{z}^k) )_l ),
\end{align}
where $k$ counts iterations and $l$ counts landmarks. $\mathcal{M}^*_l \in \mathbb{R}^{w \times h \times 1}$ is the ground truth of the response map for the $l$-th landmark. As shown in Figure~\ref{fig:fig_fsrnn}, our recurrent model progressively improves the prediction accuracy when a face exhibits challenging pose or expression. The whole process is learned end-to-end during training.


\subsection{Temporal Recurrent Learning} \label{sec:temporal}

The recurrent learning is performed at both the spatial and temporal dimensions. Given $T$ successive frames $\{\mathbf{x}^{t}; t=1, \cdots, T\}$, the encoder extracts a sequence of representation codes $\{\mathcal{C}^{t}; t=1, \cdots, T\}$. We can decouple $\mathcal{C}$ as: identity code $\mathcal{C}_{id}$ that is {\em temporal-invariant} since all frames are subject to the same identity constraint; and pose/expression code $\mathcal{C}_{pe}$ that is \textit{temporal-variant} since pose and expression changes over time \cite{PengCVIU15}. We exploit the temporal consistence of $\mathcal{C}_{pe}$ via the proposed temporal recurrent learning.

Figure \ref{fig:fig_ftrnn} shows the unrolled illustration of the proposed temporal recurrent learning. More specifically, we aim to achieve a nonlinear mapping $f_{tRNN}$, which simultaneously tracks the latent state $\{h^t;t=1,\cdots,T\}$ and updates $\mathcal{C}_{pe}$ at time $t$:
\begin{align}
	h^t = p(\mathcal{C}_{pe}^t, h^{t-1}; \theta_{tRNN}), \; {\mathcal{C}_{pe}^t}^{\prime} = q(h^t; \theta_{tRNN}), \; t=1,\cdots,T
\end{align}
where $p(\cdot)$ and $q(\cdot)$ are functions of $f_{tRNN}$. ${\mathcal{C}_{pe}^t}^{\prime}$ is the update of $\mathcal{C}_{pe}^t$. $\theta_{tRNN}$ corresponds to mapping parameters which are learned in the end-to-end task using the same classification loss as Equation \ref{eq:srnnloss} but unrolled at the temporal dimension:
\begin{align} \label{eq:trnnloss}
	\argmin_{\theta_{ENC},\theta_{DENC},\theta_{tRNN}} \sum_{t=1}^{T} \sum_{l=0}^{L} \ell_{tRNN} ( {\mathcal{M}^t_l}^*, f_{DENC}( \mathcal{C}_{id}^t, \mathcal{C}_{pe}^t )_l ),
\end{align}
where $t$ counts time steps and $l$ counts landmarks. Note that both spatial and temporal recurrent learning are performed to jointly learn $\theta_{ENC}$, $\theta_{DENC}$ and $\theta_{tRNN}$ in the same task according to Equation \ref{eq:srnnloss} and \ref{eq:trnnloss}.

The temporal recurrent learning memorize the motion patterns of pose and expression variations from offline training data. It can significantly improve the fitting accuracy and robustness when large variations and partial occlusions exist.

\subsection{Supervised Identity Disentangling} \label{sec:identity}
There is no guarantee that temporal-invariant and -variant factors can be completely decoupled in the bottleneck by simply splitting the representation codes into two parts. More supervised information is required to achieve the decoupling. To address this issue, we propose to apply a face recognition task on the identity code, in addition to the temporal recurrent learning applied on pose/expression code.

The supervised identity disentangling is formulated as an $N$-way classification problem. $N$ is the number of unique individuals present in the training sequences. In general, the classification network $f_{CLS}$ associates the identity code $\mathcal{C}_{id}$ with a vector indicating the score of each identity. Classification loss is used to learn the mapping parameters:
\begin{align}
	\argmin_{\theta_{CLS}} \sum_{m=1}^{M} \ell_{CLS} ( \mathbf{e}^*, f_{CLS}( \mathcal{C}_{id}; \theta_{CLS} ) ),
\end{align}
where $m$ counts the number of training images in a mini batch. 
$\mathbf{e}^*$ is the one-hot identity annotation vector with a $1$ for the correct identity and all $0$s for others.

It has been shown in \cite{ZhangTangECCV14} that learning the face alignment task together with correlated tasks, \textit{e.g.} head pose, can improve the fitting performance. We have the similar observation when adding face recognition task to the alignment task. More specifically, we found that supervised identity disentangling can significantly improve the generalization as well as fitting accuracy at test time. In this case, the factors are better decoupled, which facilitates $f_{tRNN}$ to better handle temporal variations.

\subsection{Constrained Shape Prediction} \label{sec:shape}

The response map output by the encoder-decoder may have a few false high responses when distractions exist in the background. Although this issue is significantly alleviated by spatial recurrent learning, it still impairs the fitting accuracy in challenging conditions. Besides, the response map uses separate channels to depict each landmark. The spatial dependencies among landmarks are not well explored. To overcome these limitations, we append nonlinear mappings after the encoder-decoder to learn the shape constraint for shape prediction.

$f_{REG}$ takes the response map as the input and outputs landmark coordinates $\mathbf{y} \in \mathbb{R}^{2L \times 1}$. Regression loss is used to learn the mapping parameters:
\begin{align}
	\argmin_{\theta_{REG}} \sum_{n=1}^{N} \ell_{REG} ( \mathbf{y}^*, f_{REG}( \mathcal{M}; \theta_{REG} ) ),
\end{align}
where $\mathbf{y}^*$ is the ground truth of landmark coordinates. All coordinates are normalized by subtracting a mean shape calculated from training images. The summation accumulates loss within a mini batch to avoid gradient jiggling.

\section{Network Architecture and Implementation Details} \label{sec:arch}
All modules are embedded in a unified framework that can be trained end-to-end. Next we provide more details about how we guarantee efficient training convergence and robust performance at test time.

\subsection{$f_{ENC}$ and $f_{DENC}$}
\begin{figure*}[t]
\centering
\includegraphics[width=0.95\textwidth]{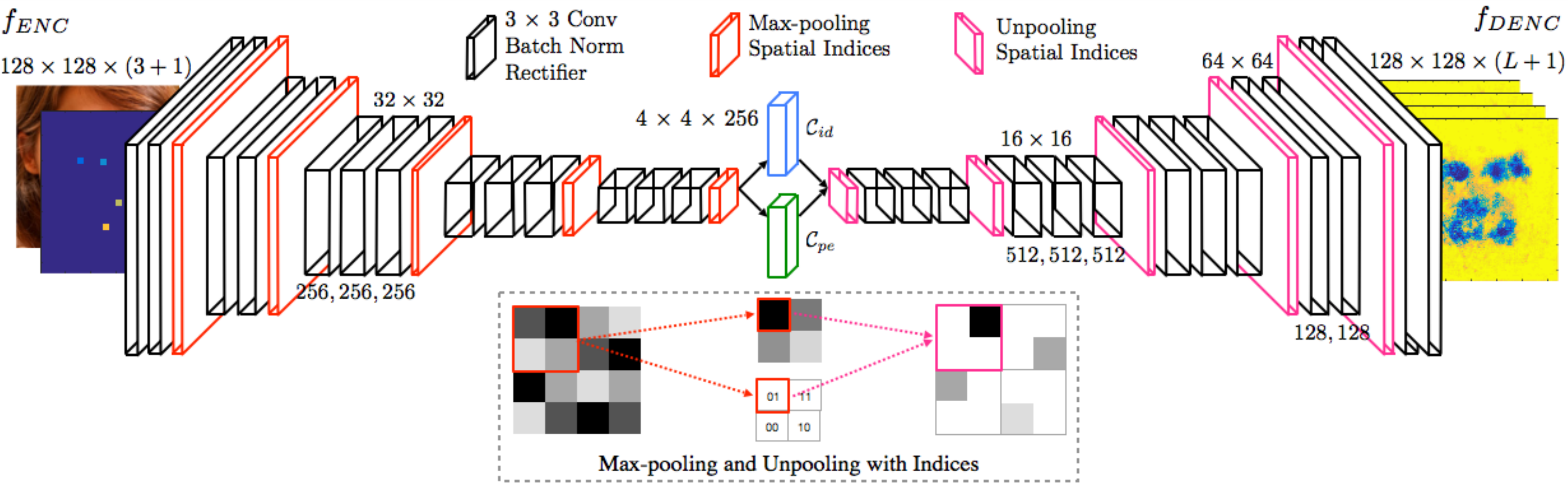}
\caption{Architecture of $f_{ENC}$ and $f_{DENC}$. The input of the encoder is the concatenation of 3-channel image and 1-channel label map. The decoder is exactly symmetrical to the encoder except the output is a $(L+1)$-channel response map. The representation code is split into $\mathcal{C}_{id}$ and $\mathcal{C}_{pe}$ in the bottleneck, where each one is a $4 \times 4 \times 256$ feature map. 3 $\times$ 3 kernels are used in all convolutional layers. 2 $\times$ 2 max-pooling or unpooling windows are applied in all pooling layers. The corresponding max-pooling and unpooling share pooling indices with a 2-bit switch for each 2 $\times$ 2 pooling window.}
\label{fig:fig_architecture1}
\end{figure*}

Figure \ref{fig:fig_architecture1} illustrates the detailed configuration of the encoder-decoder. The encoder is designed based on a variant of the VGG-16 network \cite{SimonyanCoRR14,KendallCoRR15}. It has 13 convolutional layers with constant $3 \times 3$ filters which correspond to the first 13 convolutional layers in VGG-16. We can therefore initialize the training process from weights trained on large datasets for object classification. We remove all fully connected layers  in favor of \textit{fully convolutional networks} (FCNs) \cite{LongCoRR14} and output two $4 \times 4 \times 256$ feature maps in the bottleneck. This strategy not only reduces the number of parameters from 117M to 14.8M \cite{BadriCoRR15}, but also preserves spatial information in high-resolution feature maps instead of fully-connected feature vectors, which is crucial for our landmark localization task. 

There are 5 max-pooling layers with $2 \times 2$ pooling windows and a constant stride of 2 in the encoder to halve the resolution of feature maps after each convolutional stage. Although max-pooling can help to achieve translation invariance, it inevitably results in a considerable loss of spatial information especially when several max-pooling layers are applied in succession. To solve this issue, we use a 2-bit code to record the index of the maximum activation selected in a $2 \times 2$ pooling window \cite{ZeilerECCV14}. As illustrated in Figure \ref{fig:fig_architecture1}, the memorized index is then used in the corresponding unpooling layer to place each activation back to its original location. This strategy is particularly useful for the decoder to recover the input structure from the highly compressed feature map. Besides, it is much more efficient to store the spatial indices than to memorize the entire feature map in float precision as proposed in FCNs \cite{LongCoRR14}.

The decoder is symmetrical to the encoder with a mirrored configuration but replacing all max-pooling layers with corresponding unpooling layers. The final output of the decoder is a $(L + 1)$-channel response map which is fed to a softmax classifier to predict pixel-wise confidence. We find that batch normalization \cite{IoffeCoRR15} can significantly boost the training speed as it can effectively reduce internal shift within a mini batch. Therefore, batch normalization and rectified linear unit (ReLU) \cite{NairICML10} are applied after each convolutional layer.

\begin{figure*}[t]
\centering
\includegraphics[width=0.95\textwidth]{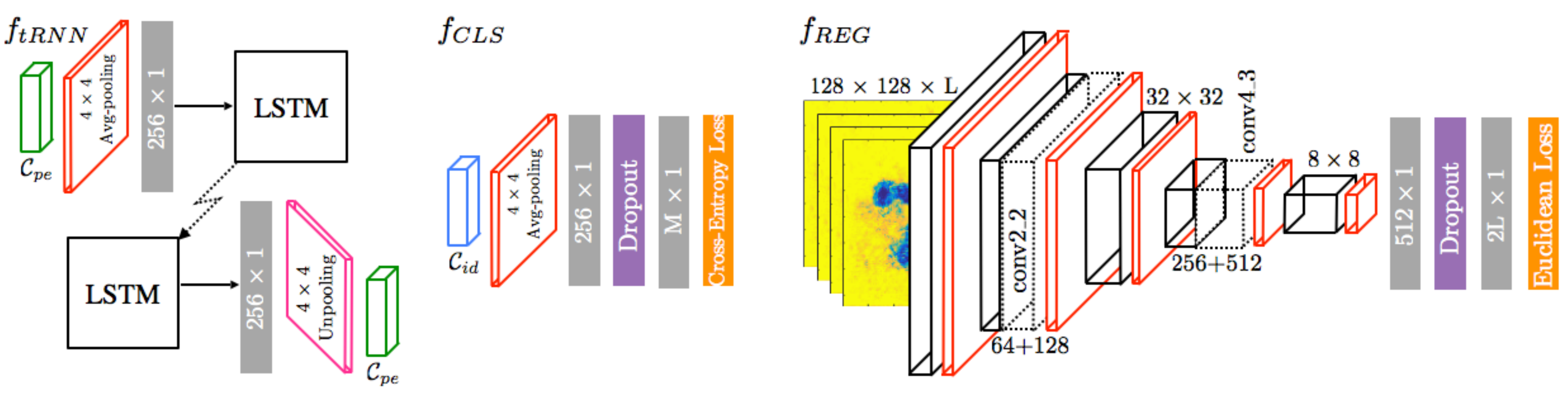}
\caption{Architecture of $f_{tRNN}$, $f_{CLS}$ and $f_{REG}$. In $f_{tRNN}$, pooling and unpooling with spatial indices are applied to cut down the input and output complexity of LSTM module. In $f_{REG}$, intermediate feature maps from the encoder, \textit{i.e.} conv2\_2 and conv4\_3, are concatenated to incorporate both global and local features.}
\label{fig:fig_architecture2}
\end{figure*}

\subsection{$f_{sRNN}$ and $f_{tRNN}$}

As shown in Figure \ref{fig:fig_overview} and \ref{fig:fig_fsrnn}, $f_{sRNN}$ maps the $(L+1)$-channel response map $\mathcal{M}$ to a single-channel label map $\mathbf{z}$. This mapping can be achieved efficiently in two steps. First, we merge $\mathcal{M}$ to a single map with $(L+1)$ clusters. The value of the map at location $(i,j)$ is set to the channel index of $\mathcal{M}$ that has the largest confidence: 
\begin{align}
	m_{ij} = \argmax_{l} (\mathcal{M}_{ij})_{l}, \ where \ l = 0,\cdots,L. 
\end{align}
The second step is to generate a label map from the clustering. We label each landmark with a small square centered at the corresponding clustering center with varied sizes. The sizes are set to 7-pixel, 5-pixel, and 3-pixel for the three recurrent steps, respectively, in order to provide the spatial feedback in a coarse-to-fine manner.

We employ Long Short-Term Memory (LSTM) \cite{HochreiterNC97,OhNIPS15} networks to model $f_{tRNN}$. $256$ hidden units are used in the LSTM layer we empirically set $T=10$. The prediction loss is calculated at each time step and then accumulated after $T$ steps for backpropagation. Directly feeding $\mathcal{C}_{pe}^t$ into the LSTM layer leads to a low training rate as it needs $4 \times 4 \times 256 = 4096$ neurons for both the input and output. We apply $4 \times 4$ pooling and unpooling to compress $\mathcal{C}_{pe}$ to a $256 \times 1$ vector as illustrated in Figure \ref{fig:fig_architecture2}.

\subsection{$f_{CLS}$ and $f_{REG}$}
To facilitate the decoupling in the bottleneck, we use a classification network to predict identity labels from $\mathcal{C}_{id}$. $f_{CLS}$ takes $\mathcal{C}_{id}$ as input and applies $4 \times 4$ average pooling to obtain a $256d$ feature vector for identity representation. Instead of using a very long feature vector in former face recognition network \cite{TaigmanCVPR14}, {\em e.g.} $4096d$, we use a more compact vector, {\em e.g.} $256d$, to reduce the computational cost without losing recognition accuracy \cite{SchroffCVPR15,SunCVPR15}. To avoid overfitting, $0.4$ dropout is applied, followed with a fully connected layer with $M$ neurons to predict the entity using the cross-entropy loss. 

The regression network takes $128 \times 128 \times (L+1)$ response map as input to directly predict $2L \times 1$ normalized landmark coordinates. The network architecture is similar to the encoder but using fewer feature maps in each convolutional layer: 64-64-256-256-512. The dimension of feature maps is halved after each $2 \times 2$ max-pooling layer except the last $8 \times 8$ pooling layer to achieve a $512d$ feature vector. Similar to the classification network, $0.4$ dropout is applied. A fully connected layer with $2L \times 1$ neurons is used to output landmark coordinates, which is used to compute the Euclidean loss.

We experienced suboptimal performance with the designed $f_{REG}$ at the beginning. The reason is the response map is highly abstract and missing detailed information of the input image. To address this issue, we incorporate feature maps from the encoder to boost the regression accuracy. More specifically, we concatenate feature maps from both shallow layer ({\em conv2\_2}) and deep layer ({\em conv4\_3}) to the corresponding layers in $f_{REG}$ to utilize both global and local features. Figure \ref{fig:fig_architecture2} illustrates the idea. 
Both {\em conv2\_2} and {\em conv4\_3} are learned in the encoder-decoder and remain unchanged in $f_{REG}$.

\section{Experiments}

In this section, we first demonstrate the effectiveness of each component in our framework, followed with performance comparison against the state-of-the-arts on both controlled and unconstrained datasets.

\subsection{Datasets and Settings}
\textbf{Datasets.} We conduct our experiments on widely used benchmark datasets as listed in Table \ref{tab:data}. These datasets present challenges in multiple aspects such as large pose, extensive expression variation, severe occlusion and dynamic illumination. 

We generated 7-landmark annotation for all datasets to locate eye corners, nose tip and mouth corners. Besides, we followed \cite{SagonasICCVW13} for unified 68-landmark annotation for Helen, LFPW, Talking Face (TF), Face Movie (FM) and 300-VW. Moreover, we manually labeled the identity for each video in TF, FM and 300-VW. The landmark annotation of LFW is given by \cite{Gary14}. 

AFLW and 300-VW have the largest number of labeled images. They are also more challenging than others due to the extensive variations. Therefore, we used them for both training and evaluation. More specifically, $80\%$ of the images in AFLW and $90$ out of $114$ videos in 300-VW were used for training, and the rest were used for evaluation. We sampled videos to roughly cover the three different scenarios defined in \cite{ChrysosICCVW15}, \textit{i.e.} "Scenario 1", "Scenario 2" and "Scenario 3", corresponding to well-lit, mild unconstrained and completely unconstrained conditions, respectively.

We performed data augmentation by sampling ten variations from each image in the image training datasets. The sampling was achieved by random perturbation of scale ($0.9$ to $1.1$), rotation ($\pm 15^\circ$), translation ($7$ pixels), as well as horizontal flip. To generate sequential training data, we randomly sampled 100 clips from each training video, where each clip has 10 frames. It is worthy mentioning that no augmentation is applied on video training data to preserve the temporal consistency in the successive frames.

\begin{table*}[t]
\centering
\caption{The \textit{image} and \textit{video datasets} used in training and evaluation. LFW, TF, FM and 300-VW have both landmark and identity annotation. AFLW and 300-VW are split into two sets for both training and evaluation.} \label{tab:data}
\begin{tabular}{c | c c c c | c  c c }
\toprule
& AFLW \cite{Koestinger11}  & LFW \cite{Gary14} & Helen \cite{LeECCV12} & LFPW \cite{BelhumeurCVPR11} & TF \cite{fgnet04} & FM \cite{PengICCV15} & 300-VW \cite{ShenICCVW15} \\
\hline
in-the-wild & Y & Y & Y & Y & N & Y & Y \\
image  & 21,080 & 12,007 & 2,330 & 1,035 & 500 & 2,150 & 114,000 \\
video  & - & - & - & - & 5 & 6 & 114 \\ 
landmark  & 21 & 7 & 194 & 68 & 68 & 68 & 68 \\
identity  & - & 5,371 & - & - & 1 & 6 & 105 \\
\hline
training  & 16,864 & 12,007 & 2,330 & 1,035 & 0 & 0 & 90,000 \\
evaluation  & 4,216 & 0 & 0 & 0 &  500 & 2150 & 24,000 \\
\bottomrule
\end{tabular}
\end{table*}

\noindent \textbf{Training.} Our approach is capable of end-to-end training on the video datasets. However, there are only 105 different identities in 300-VW. 
To make full use of all annotated datasets, we conducted the training through three steps. In each step, we optimized the network parameters by using \textit{stochastic gradient descent} (SGD) with 0.9 momentum. The learning rate started at 0.01 and decayed $20\%$ after every 10 epochs. 

In the first step, we trained the network without $f_{CLS}$ and $f_{tRNN}$ using AFLW, Helen and LFPW. We initialized $f_{ENC}$ using pre-trained weights in VGG-16 \cite{SimonyanCoRR14}, and left other modules with Gaussian initialization \cite{JiaACMM14}. The training was performed for 30 epochs. In the second step, we added $f_{CLS}$ and fine-tuned other modules using LFW. The training was performed for 20 epochs. In the third step, we added $f_{tRNN}$ and fine-tuned the entire network using 300-VW. The mini-batch size was set to 5 clips that had no identity overlap to avoid oscillations of the identity loss. For each training clip, we performed temporal recurrent learning for another 50 epochs in both forward and backward direction to double the training data.

\noindent \textbf{Evaluation.} To avoid overfitting, we ensure that the training and testing videos do not have identity overlap on the 300-VW (16 videos share 7 identities). We used normalized \textit{root mean square error} (RMSE) \cite{SagonasICCVW13} for fitting accuracy evaluation.  A prediction with larger than $10\%$ mean error was reported as a failure \cite{ShenICCVW15,TangTIP12}.

\subsection{Validation of Spatial Recurrent Learning}

We validate the proposed spatial recurrent learning on the validation set of AFLW. To better investigate the benefits of spatial recurrent learning, we partitioned the validation set into four image groups according to the absolute value of yaw angle \cite{PengFG15}: $0^{\circ}$-$15^{\circ}$, $15^{\circ}$-$30^{\circ}$, $30^{\circ}$-$45^{\circ}$ and $45^{\circ}$-$90^{\circ}$.

\begin{figure}[t]
\minipage{0.49\textwidth}
\center
  \includegraphics[width=.98\linewidth]{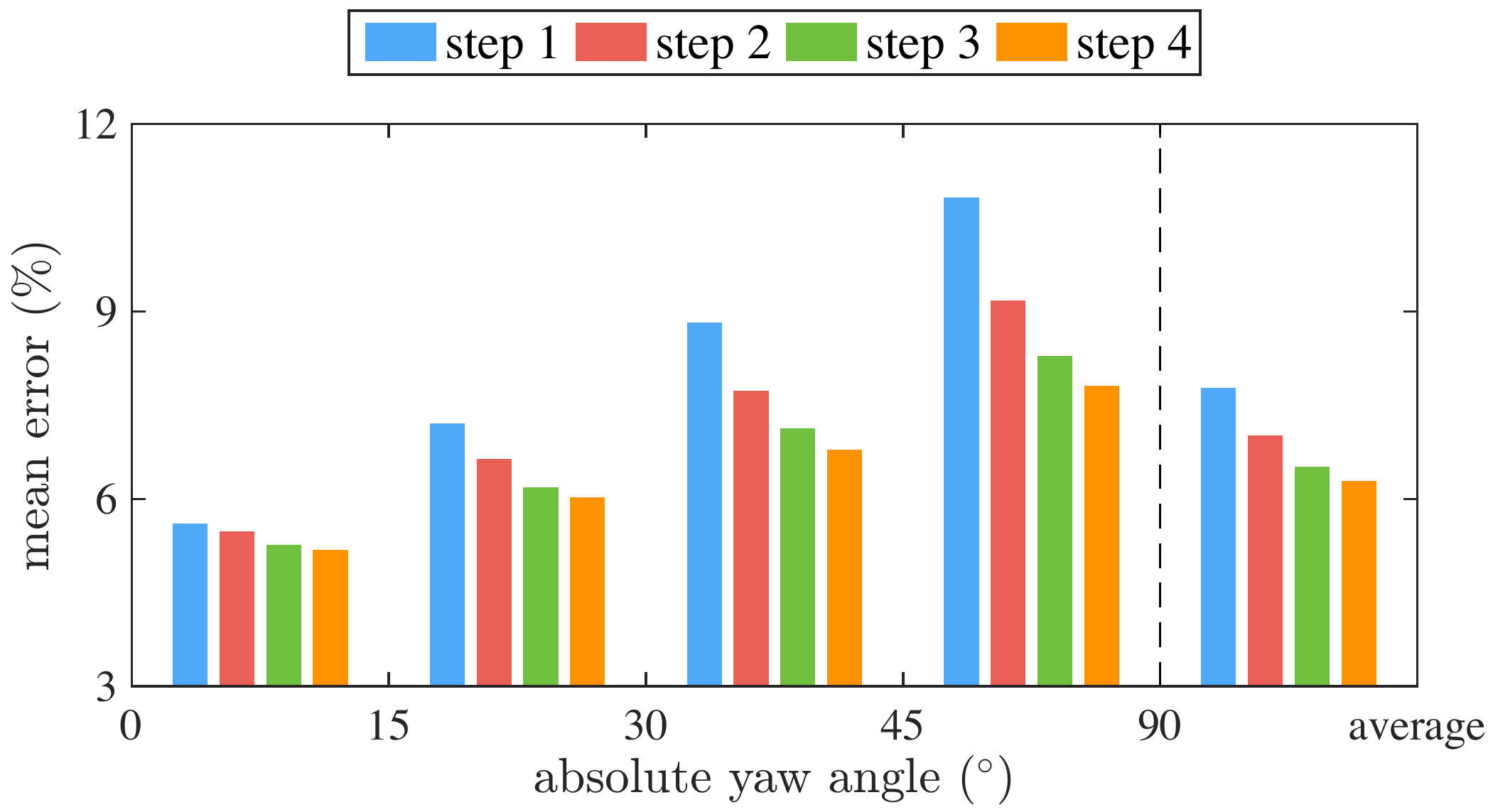}
  \caption{Mean errors after each spatial recurrent step on the validation set of AFLW \cite{Koestinger11}. The fitting improvement is more significant on faces with large head poses ($45^{\circ}$-$90^{\circ}$) than near frontal faces ($0^{\circ}$-$15^{\circ}$). Three-step recurrent learning achieve a good trade-off between fitting accuracy and efficiency, as the fourth step has very limited improvement.}\label{fig:fig_fsrnn_hist}
\endminipage\hfill
\minipage{0.49\textwidth}
\center
  \includegraphics[width=0.94\linewidth]{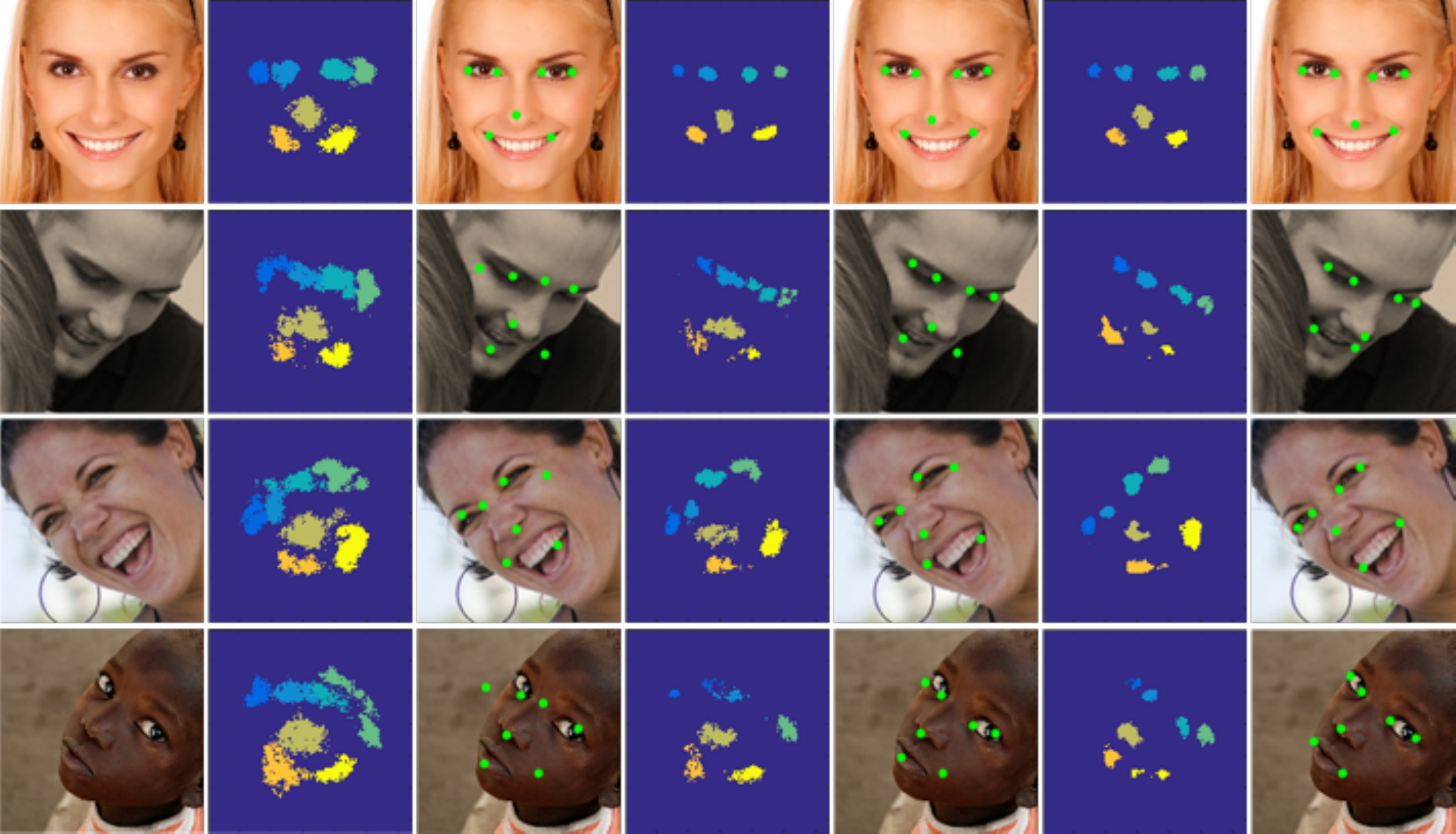}
  \caption{Examples of three-step spatial recurrent learning. Successive recurrent steps are not necessary in easy cases (first row), but is crucial in challenging cases such as large pose and intense expression (rest of rows). The response clusters shrink and converge in successive recurrent steps, which moves landmarks toward ground truth step by step.}\label{fig:fig_fsrnn_img}
\endminipage\hfill
\end{figure}

First, we trained a 4-step recurrent model and reported the mean error after each step in Figure \ref{fig:fig_fsrnn_hist}. From which, we had the following observations: \textbf{(1)} The fitting errors decrease in the successive recurrent steps. \textbf{(2)} The improvement of fitting accuracy is much more significant on faces with large head poses than near frontal faces, \textit{e.g.} $23.3\%$ improvement on $45^{\circ}$-$90^{\circ}$ set and $6.10\%$ improvement on $0^{\circ}$-$15^{\circ}$ pose set. \textbf{(3)} The improvement is saturated after the first three recurrent steps as the fourth step has very limited improvement. These observations validate the proposed spatial recurrent learning to improve the fitting accuracy especially in challenging cases such as large pose. Besides, we set the number of recurrent steps to 3 in the following experiments, as it achieves a good trade-off between fitting accuracy and efficiency. Figure \ref{fig:fig_fsrnn_img} shows examples of recurrent learning. The response clusters shrink and converge in successive recurrent steps, which moves landmarks from initial to ground truth step by step. 

\begin{table*}[t]
\centering
\caption{Mean error comparison between the proposed spatial recurrent learning and the widely used cascade learning on large pose ($> 30^{\circ}$) set of AFLW. Each network in cascade has exactly the same architecture as the recurrent version but not sharing weight among cascades. The recurrent learning beats the cascade variant in terms of fitting accuracy and efficiency.} \label{tab:cascadevsrecurrent}
\begin{tabular}{c | C{1.2cm} C{1.2cm} C{1.cm} C{1.cm} | C{1.cm} C{1.cm} C{1.cm} | C{1.cm} C{1.2cm}}
\toprule
& Left eye  & Right eye  & Nose  & Mouth & Mean  & Std & Failure  & Time & Memory \\
& $\%$ & $\%$ & $\%$ & $\%$ & $\%$ & $\%$ & $\%$ & ms & MB \\
\hline
Cascade		& 8.32& 7.37 & 9.60 & 7.52 & 8.07 & 6.36 & 25.5  & 31.7 & 88.9 \\ 
Recurrent		& 7.59 & 7.21 & 8.76 & 6.45 & 7.33 & 3.94 & 17.3  & 28.5 & 29.6 \\
\bottomrule
\end{tabular}
\end{table*}

Second, it is reasonable to compare the proposed spatial recurrent learning with the widely used cascade learning such as \cite{SunCVPR13,ZhangECCV14}. For a fair comparison, we implemented a three-step cascade variant of our approach. Each network in the cascade has exactly the same architecture as the spatial recurrent version but there is no weight sharing among cascades. We fully trained the cascade networks using the same training set and validated the performance on the large pose (> $30^\circ$) set of AFLW. The comparison is presented in Table \ref{tab:cascadevsrecurrent}. We can see that the spatial recurrent learning can significantly improve the fitting performance. The underlying reason is the recurrent network learns the tep-by-step fitting strategy jointly, while the cascade networks learn each step independently. It can better handle the challenging case where the initial guess is usually far away from the ground truth. Moreover, a single network with shared weights can instantly reduce the memory usage to one third of the cascaded implementation. 

\subsection{Validation of Temporal Recurrent Learning}
In this section, we validate the proposed temporal recurrent learning on the validation set of 300-VW. To better study the performance under different settings, we split the validation set into two groups: 9 videos in common settings that roughly match "Scenario 1", and 15 videos in challenging settings that roughly match "Scenario 2" and "Scenario 3". The common, challenging and full sets were used in the following evaluation.

We implemented a variant of our approach that turns off the temporal recurrent learning $f_{tRNN}$. It was also pre-trained on the image training set and fine-tuned on the video training set. Since there was no temporal recurrent learning, we used frames instead of clips to conduct the fine-tuning which was performed for the same 50 epochs. We showed the result with and without temporal recurrent learning in Table \ref{tab:ftrnn}.

\begin{table*}[t]
\centering
\caption{Mean error comparison between the proposed temporal recurrent learning and the variant without $f_{tRNN}$ on the validation set of 300-VW \cite{SagonasICCVW13}. The temporal recurrent learning significantly improves the tracking accuracy (smaller mean error) and robustness (smaller std and lower failure rate), especially on the validation set in challenging settings.  } \label{tab:ftrnn}.
\begin{tabular}{ c  C{1cm} C{1cm} C{1cm} c C{1cm} C{1cm} C{1cm} c C{1cm} C{1cm} C{1cm} }
\toprule
& \multicolumn{3}{c}{Common} & & \multicolumn{3}{c}{Challenging} & & \multicolumn{3}{c}{Full}\\
\cline{2-4} \cline{6-8} \cline{10-12}
& Mean & Std & Failure & & Mean & Std & Failure & & Mean & Std & Failure\\
& $\%$ & $\%$ & $\%$ & & $\%$ & $\%$ & $\%$ & & $\%$ & $\%$ & $\%$\\
\hline
w/o $f_{tRNN}$ 	& 4.52 & 2.24 & 3.48 &  & 6.27 & 5.33 & 13.3 &  & 5.83 & 3.42 & 6.43 \\
$f_{tRNN}$ 			& 4.21 & 1.85 & 1.71 &  & 5.64 & 3.28 & 5.40 &  & 5.25 & 2.15 & 2.82\\
\bottomrule
\end{tabular}
\end{table*}

\begin{figure*}[t]
\centering
\includegraphics[width=0.95\textwidth]{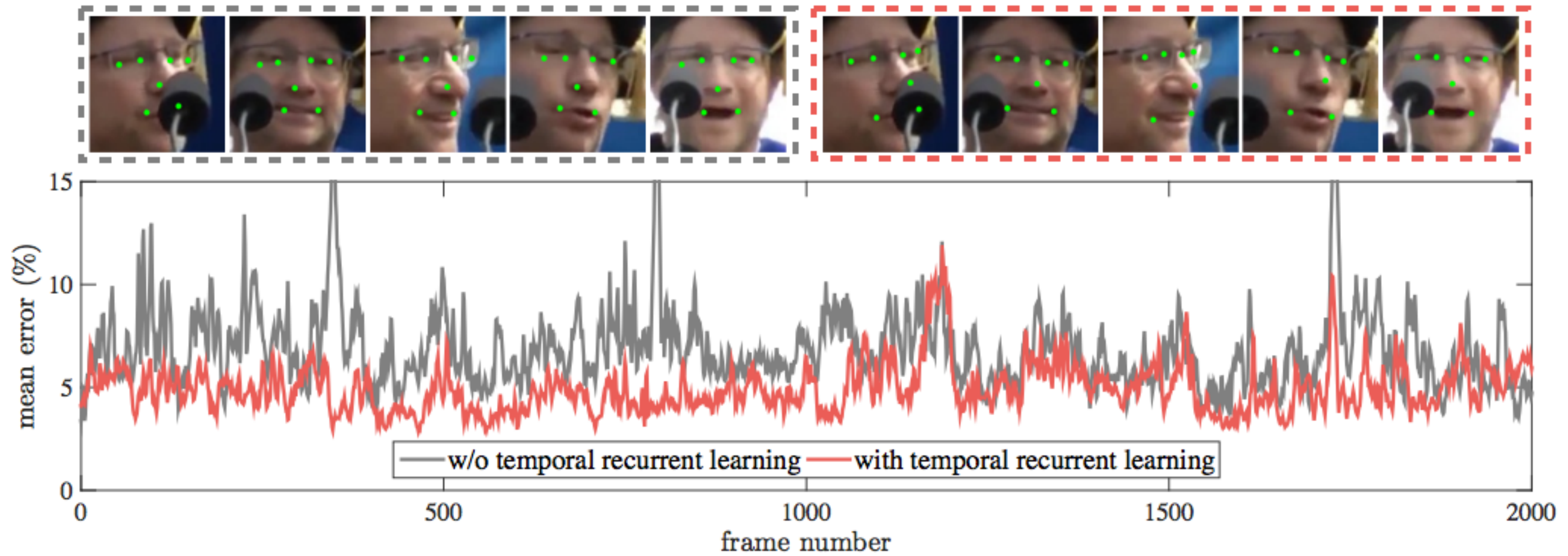}
\caption{Examples of validation results in challenging settings. The tracked subject undergoes intensive pose and expression variations as well as severe partial occlusions. The proposed temporal recurrent learning has substantial improvement in terms of tracking accuracy and robustness, especially for landmarks on nose tips and mouth corners.} \label{fig:fig_ftrnn_curve}
\end{figure*}

For videos in common settings, the temporal recurrent learning achieves $6.8\%$ and $17.4\%$ improvement in terms of mean error and standard deviation respectively, while the failure rate is remarkably reduced by $50.8\%$. Temporal modeling produces better prediction by taking consideration of history observations. It may implicitly learn to model the motion dynamics in the hidden units from the training clips. 

For videos in challenging settings, the temporal recurrent learning won with even bigger margin. Without $f_{tRNN}$, it is hard to capture the drastic motion or changes in consecutive frames, which inevitably results in higher mean error, std and failure rate. Figure \ref{fig:fig_ftrnn_curve} shows an example where the subject exhibits intensive pose and expression variations as well as severe partial occlusions. The curve showed our recurrent model obviously reduced landmark errors, especially for landmarks on nose tip and mouth corners. The less oscillating error also suggests that $f_{tRNN}$ significantly improves the prediction stability over frames.

\subsection{Benefits of Supervised Identity Disentangling}

The supervised identity disentangling is proposed to better decouple the temporal-invariant and temporal-variant factors in the bottleneck of the encoder-decoder. This facilitates the temporal recurrent training, yielding better generalization and more accurate fittings at test time.

\begin{figure*}[t]
\centering
\subfloat{\includegraphics[width=.47\linewidth]{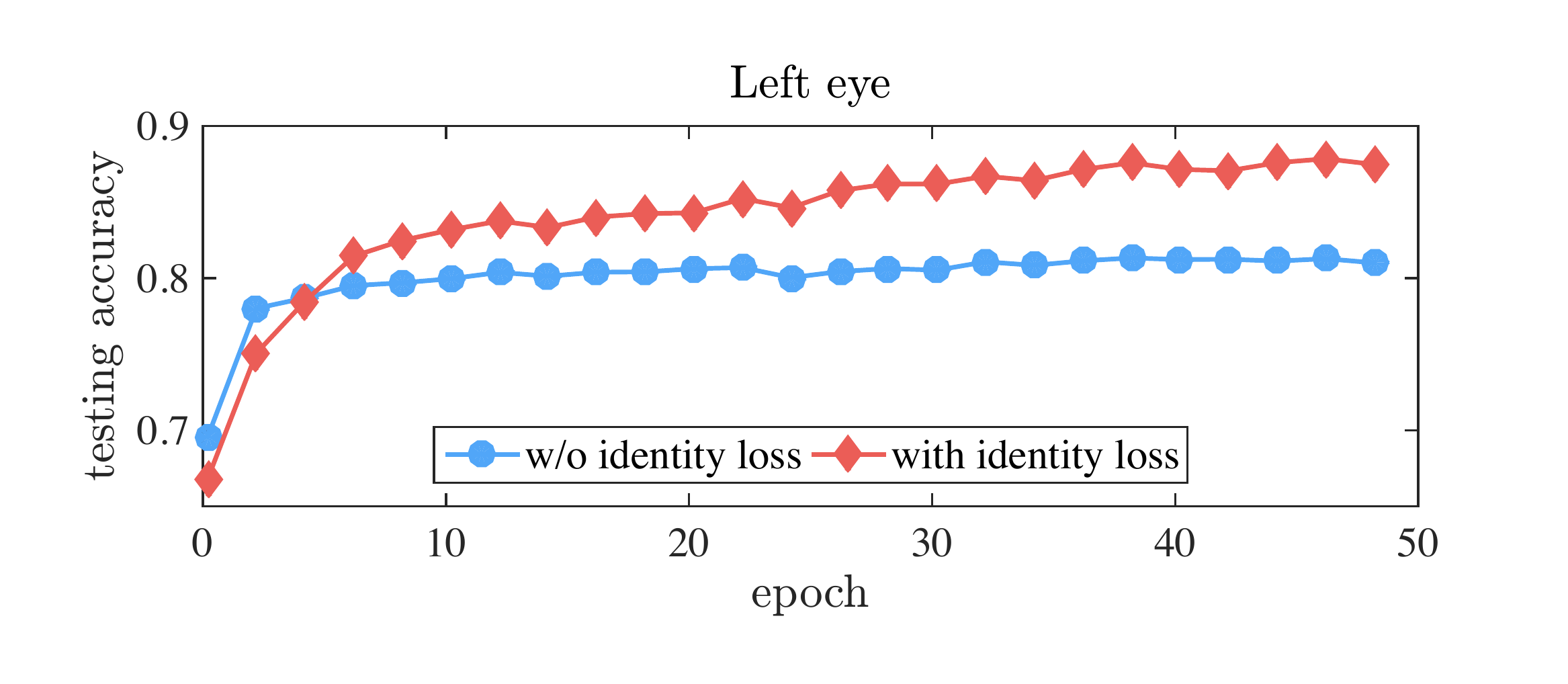}}\ \subfloat{\includegraphics[width=.47\linewidth]{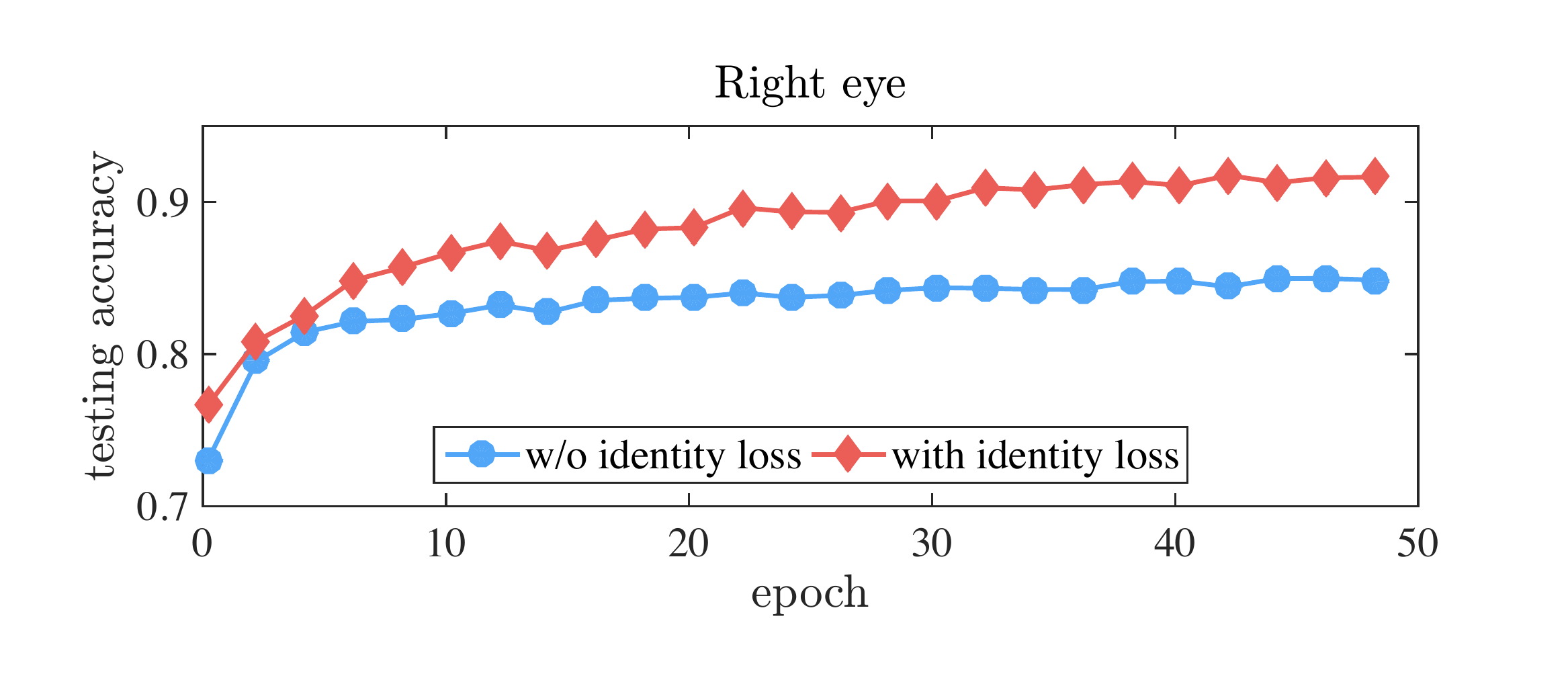}}\\
 \subfloat{\includegraphics[width=.47\linewidth]{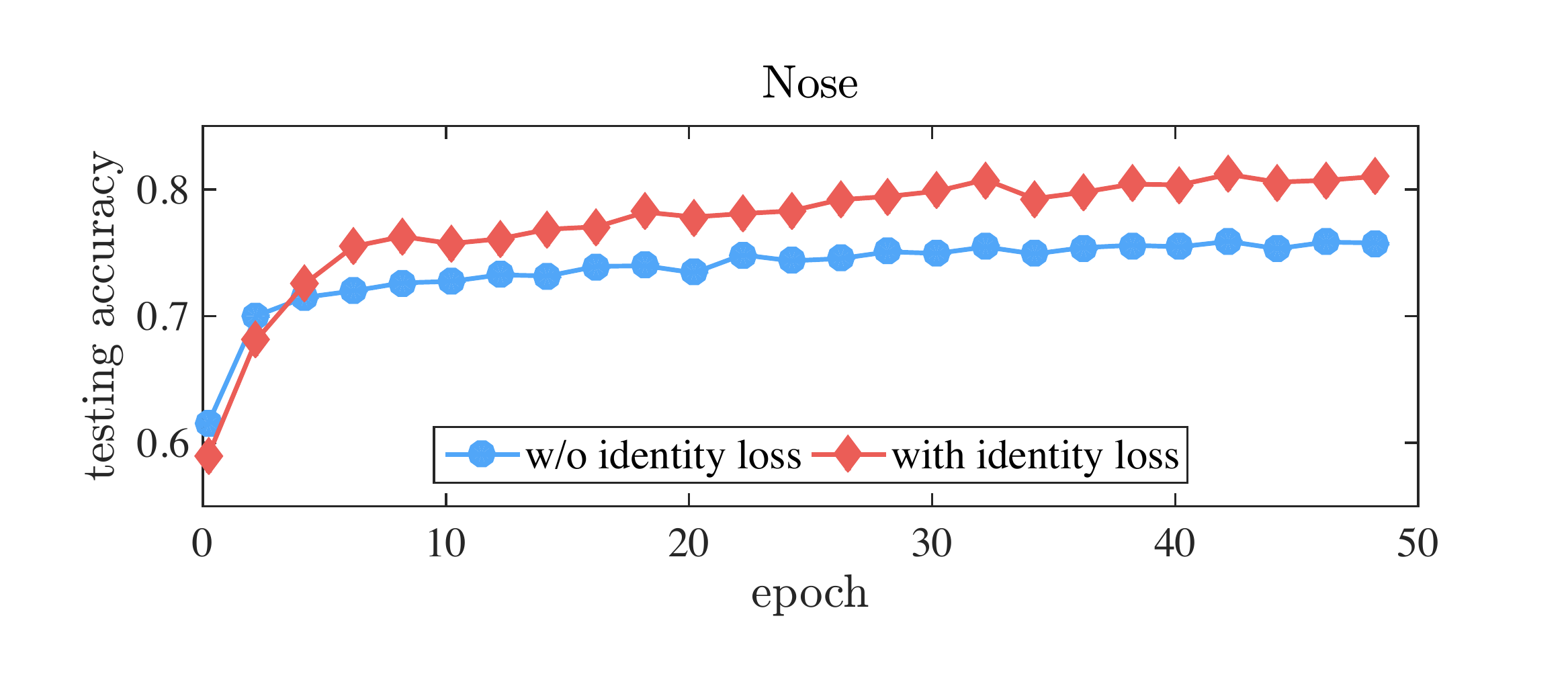}}\ \subfloat{\includegraphics[width=.47\linewidth]{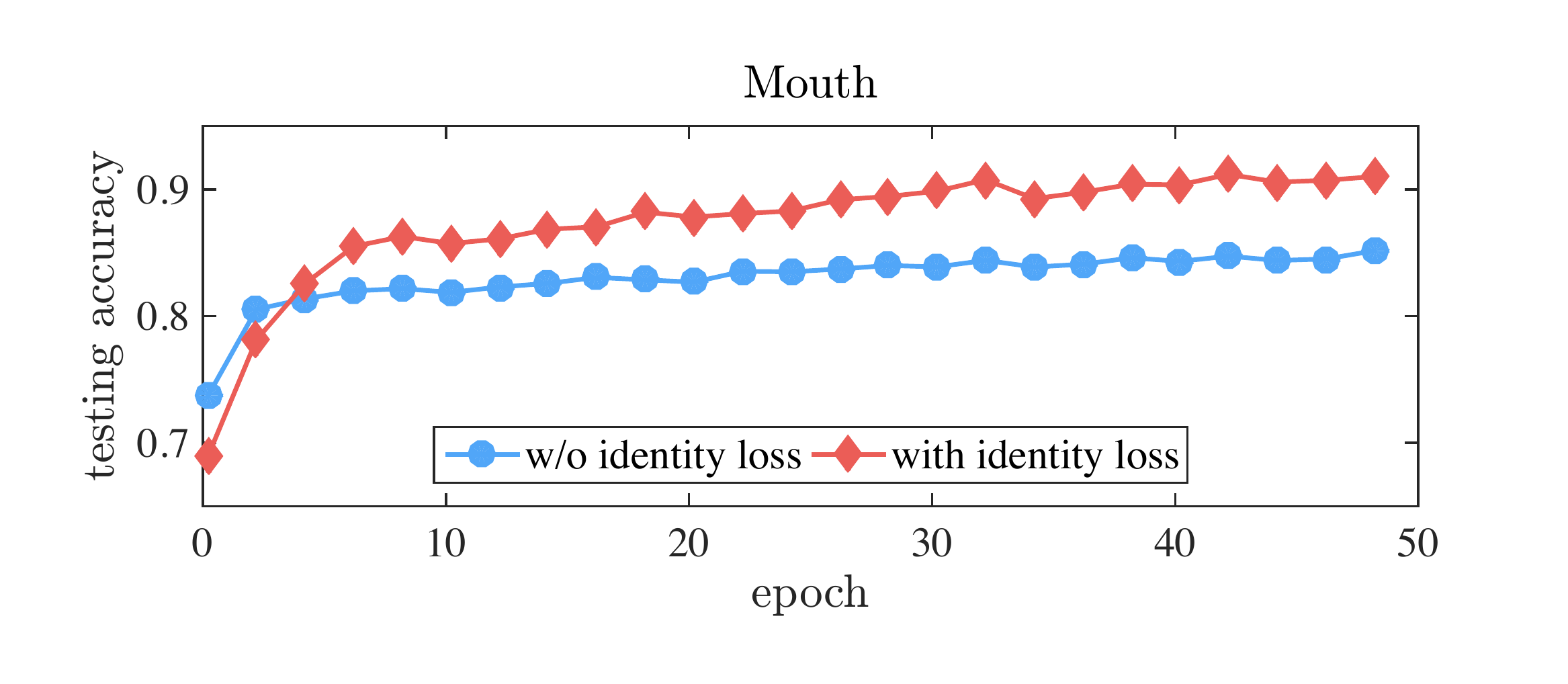}} 
  \caption{Testing accuracy of different facial components with respect to the number of training epochs. The proposed supervised identity disentangling helps to achieve a more complete factor decoupling in the bottleneck of the encoder-decoder, which yields better generalization capability and more accurate testing results.}\label{fig:fig_fcls}
\end{figure*}

To study the effectiveness of the identity network,   we removed $f_{CLS}$ and follow the exact training steps. The testing accuracy comparison on the 300-VW dataset is shown in  Figure \ref{fig:fig_fcls}. The accuracy was calculated as the ratio of pixels that were correctly classified in the corresponding channel(s) of the response map.

The validation results of different facial components show similar trends: \textbf{(1)} The network demonstrates better generalization capability by using additional identity cues, which results in a more efficient training. For instance, after only 10 training epochs, the validation accuracy for landmarks located at the left eye reaches 0.84 with identity loss compared to 0.8 without identity loss. \textbf{(2)} The supervised identity information can substantially boost the testing accuracy. There is an approximately $9\%$ improvement by using the additional identity loss. It worth mentioning that, at the very beginning of the training (< 5 epochs), the network has inferior testing accuracy with supervised identity disentangling. It is because the suddenly added identity loss perturbs the backpropagation process. However, the testing accuracy with identity loss increases rapidly and outperforms the one without identity loss after only a few more training epochs.

\subsection{Comparison with State-of-the-art methods}
We compared our framework with both traditional approaches and deep learning based approaches. The methods with hand-crafted features include: \textbf{(1)} DRMF \cite{AsthanaCVPR13}, \textbf{(2)} ESR \cite{CaoIJCV14}, \textbf{(3)} SDM \cite{XiongCVPR13}, \textbf{(4)} IFA \cite{AsthanaCVPR14}, and \textbf{(5)} PIEFA \cite{PengICCV15}. The deep learning based methods include: \textbf{(1)} DCNC  \cite{SunCVPR13}, \textbf{(2)} CFAN \cite{ZhangECCV14}, and \textbf{(3)} TCDCN \cite{ZhangTangECCV14}. All these methods were recently proposed and reported state-of-the-art performance. For fair comparison, we evaluated these methods in a tracking protocol: fitting result of current frame was used as the initial shape (DRMF, SDM and IFA) or the bounding box (ESR and PIEFA) in the next frame. The comparison was performed on both controlled, \textit{e.g.} Talking Face (TF) \cite{fgnet04}, and in-the-wild datasets, \textit{e.g.} Face Movie (FM) \cite{PengICCV15} and 300-VW \cite{ShenICCVW15}. 

\begin{table*}[t]
\centering
\caption{Mean error comparison with state-of-the-art methods on multiple video validation sets. The top performance in each dataset is highlighted. Our approach achieves the best fitting accuracy on both controlled and unconstrained datasets. } \label{tab:soa_compare}.
\begin{tabular}{ c C{3em} C{4em} C{6em} c c C{3em} C{4em} C{6em} }
\toprule
& \multicolumn{3}{c}{7 landmarks} & & & \multicolumn{3}{c}{68 landmarks} \\
\cline{2-4} \cline{7-9} 
& TF\cite{fgnet04} & FM\cite{PengICCV15} & 300-VW\cite{ShenICCVW15} & & & TF\cite{fgnet04} & FM\cite{PengICCV15} & 300VW\cite{ShenICCVW15} \\
& $\%$ & $\%$ & Challenging & & & $\%$ & $\%$ & Challenging \\
\cline{1-4} \cline{6-9} 
DRMF \cite{AsthanaCVPR13} & 4.43 & 8.53 & 9.16 & & ESR  \cite{CaoIJCV14} & 3.49 & 6.74 & 7.09\\
ESR  \cite{CaoIJCV14} & 3.81 & 7.58 & 7.83 & & SDM \cite{XiongCVPR13} & 3.80 & 7.38 & 7.25\\
SDM \cite{XiongCVPR13} & 4.01 & 7.49 & 7.65 & & CFAN \cite{ZhangECCV14} & 3.31 &6.47 & 6.64\\
IFA \cite{AsthanaCVPR14} & 3.45 & 6.39 & 6.78 & & TCDCN \cite{ZhangTangECCV14} & 3.45 & 6.92 & 7.59 \\
DCNC \cite{SunCVPR13} & 3.67 & 6.16 & 6.43 & & PIEFA \cite{PengICCV15} & 3.24 & $\bf{6.07}$ & 6.37\\
\cline{1-4} \cline{6-9} 
OURS & $\bf{3.32}$ & $\bf{5.43}$ & $\bf{5.64}$ & & OURS & $\bf{3.17}$ & 6.18 & $\bf{6.25}$\\ 
\bottomrule
\end{tabular}
\end{table*}

We report the evaluation results for both 7 and 68 landmark setups in Table ~\ref{tab:soa_compare}. Our approach achieves state-of-the-art performance under both settings. It outperforms others with a substantial margin on all datasets under 7-landmark evaluation. The performance gain is more significant on the challenging datasets (FM and 300-VW) than controlled dataset (TF). The performance of our approach degrades slightly under 68-landmark evaluation. It is a reasonable degradation considering training images (3k) that have 68-landmark annotation are much less than the ones that have 7-landmark annotation (30k). Although the training set of 300-VW contains 90k frames, the variations are limited as only 105 different identities are present. Our alignment model runs fairly fast, it takes around 30ms to process an image using a Tesla K40 GPU accelerator.

\section{Future Work}
In this paper, we proposed a novel recurrent encoder-decoder network for real-time sequential face alignment. Intensive experiments demonstrated the effectiveness of our framework and its superior performance. It decouples temporal-invariant and -variant factors in the bottleneck of the network, and exploits recurrent learning at both spatial and temporal dimensions.

The proposed method provides a general framework that can be further applied to other localization-sensitive tasks, such as human pose estimation, object detection, scene classification, etc. In the future, we plan to further exploit the proposed recurrent encoder-decoder network for boarder impact.

\bibliographystyle{splncs03}
\bibliography{egbib}
\end{document}